# Detecting and Classifying Defective Products in Images Using YOLO

Zhen Qi[1,a] , Liwei Ding[2,b], Xiangtian Li [3,c], Jiacheng Hu [4,d], Bin Lyu [5,e] , Ao Xiang [6,f]

[1] Northeastern University, Electrical and Computer Engineering, Boston, USA

[2] Florida International University, Computer Science, Miami, USA

[3] University of California San Diego, Electrical and Computer Engineering, San Diego, USA

[4] Tulane University, Business Analytics, New Orleans, USA

[5] Kanazawa University, Division of Mathematical and physical sciences, Ishikawa, Japan

[6] Northern Arizona University, Information Security and Assurance, Arizona, USA
[a] garyzhen79@gmail.com, [b] lding@fiu.edu,
[c] xil160@ucsd.edu, [d] jhu10@tulane.edu, [e] q873541969@gmail.com,
[f] ax36@nau.edu

**Abstract:** With the continuous advancement of industrial automation, product quality inspection has become increasingly important in the manufacturing process. Traditional inspection methods, which often rely on manual checks or simple machine vision techniques, suffer from low efficiency and insufficient accuracy. In recent years, deep learning technology, especially the YOLO (You Only Look Once) algorithm, has emerged as a prominent solution in the field of product defect detection due to its efficient real-time detection capabilities and excellent classification performance. This study aims to use the YOLO algorithm to detect and classify defects in product images. By constructing and training a YOLO model, we conducted experiments on multiple industrial product datasets. The results demonstrate that this method can achieve real-time detection while maintaining high detection accuracy, significantly improving the efficiency and accuracy of product quality inspection. This paper further analyzes the advantages and limitations of the YOLO algorithm in practical applications and explores future research directions.

**Keywords:** Product defect detection, YOLO, deep learning, real-time detection, industrial automation

## 1. Introduction

In modern manufacturing, product quality control is a critical component in ensuring production efficiency and customer satisfaction. With the widespread adoption of industrial automation, traditional manual inspection methods can no longer meet the increasing production demands. Manual inspection is not only time-consuming and labor-intensive but also prone to subjective factors, leading to inconsistent results. To overcome these challenges, computer vision technology has gradually been introduced into product quality inspection. However, early computer vision systems primarily relied on feature engineering-based methods, which showed clear limitations when dealing with complex scenarios and diverse product defects. In recent years, with the rapid development of deep learning technology, particularly the widespread application of convolutional neural networks (CNNs), significant progress has been made in product defect detection technology. YOLO (You Only Look Once), as a deep learning-based object detection algorithm, has gained considerable attention for its high speed and accuracy[1]. YOLO can simultaneously perform object detection and classification tasks in a single neural network forward pass, making it uniquely advantageous in real-time detection scenarios. Therefore, applying the YOLO algorithm to product defect detection can not only improve detection efficiency but also effectively enhance accuracy and consistency[2]. The primary objective of this study is to explore how to utilize the YOLO algorithm to detect and classify defects in product images in industrial production. Through experiments on multiple industrial datasets, we aim to validate the effectiveness of the YOLO algorithm in product quality inspection and analyze its feasibility and limitations in practical applications[3]. This paper will detail the working principles of the YOLO algorithm, the model training process, experimental design, and result analysis, and will discuss the practical application prospects and future research directions of this method in industrial production lines[4].

## 2. Concepts and Applications of Product Defect Detection Based on YOLO Algorithm

The YOLO (You Only Look Once) algorithm, known for its excellent real-time performance and efficiency, is widely used in the field of product defect detection. In modern manufacturing, timely and accurate detection of defects in products is crucial for maintaining the efficient operation of production lines[5]. <Figure 1> illustrates the process of product defect detection based on the YOLO algorithm, which can be divided into three main parts: input data acquisition, anomaly detection, and labeling strategy[6].

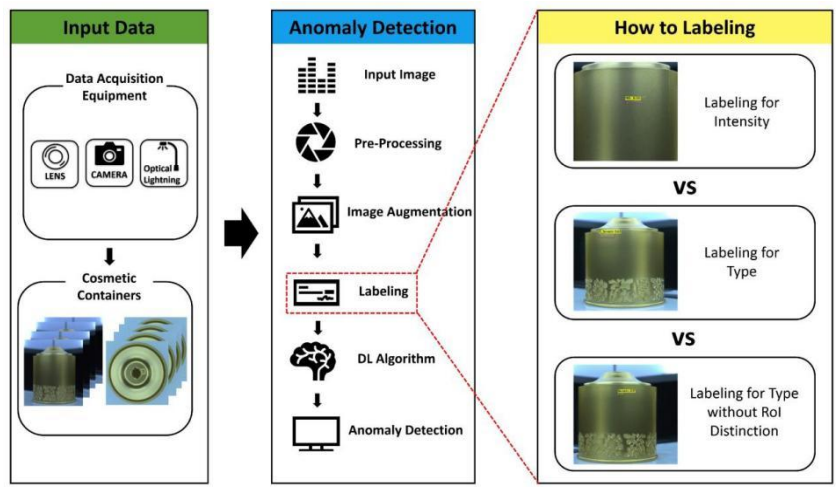

*Figure 1: Product Defect Detection Process Based on YOLO Algorithm*

First, the input data is acquired using data collection equipment such as optical lenses, cameras, and lighting devices, resulting in high-quality images containing potential product defects. In this example, the research subjects are machine parts[7]. The acquired image data undergo preprocessing and image enhancement steps to ensure that the model can extract effective features. Specifically, during preprocessing, adjustments are made to the image contrast and brightness to minimize the impact of lighting variations on detection results[8]. Next, the image data enters the anomaly detection phase. The YOLO algorithm, as the core of deep learning, is employed to extract features from the input images, detect, and classify them[9]. To ensure detection accuracy, image augmentation techniques are used to expand the dataset's diversity, thereby improving the model's generalization ability. The labeling strategy depicted in the figure is a crucial step in the detection process. Different labeling strategies during model training can significantly impact the final detection results. The right side of Figure 1 shows three different labeling methods: intensity-based labeling, type-based labeling, and type labeling without Region of Interest (RoI) distinction. These different labeling strategies allow the model to more accurately locate and identify various defects.Finally, the trained YOLO model detects defects in the input product images and outputs a conclusion on whether anomalies exist[10]. This process can run in real-time on the production line, significantly improving detection efficiency and allowing for the early identification and correction of potential defects, thereby avoiding quality issues in subsequent large-scale production[11]. The concept of product defect detection based on the YOLO algorithm has already been applied in many industrial fields. It not only significantly improves detection speed but also maintains high accuracy and stability in complex manufacturing environments. However, the YOLO algorithm still faces challenges in practical applications, such as its reliance on the diversity of datasets and the model's sensitivity to specific defect types[12]. Therefore, future research will focus on further optimizing the YOLO algorithm and labeling strategies to meet the detection needs of various complex production environments[13].

## 3. Methodology

### 3.1. Dataset Preparation

The image dataset used in this study encompasses various types of defects in different machine parts to effectively train and test the YOLO model. As shown in Table 1, the images in the dataset are sourced from industrial production lines and laboratory simulations, covering common machine parts such as bearings, gears, and bolts. All images were captured in controlled environments using high-resolution cameras and standardized lighting equipment to ensure image quality and consistency[14].

*Table 1: Overview of the Dataset*

| ataset Name | Number of Images | Resolution | Defect Types | Notes |
|---|---|---|---|---|
| Bearings (Type A) | 1200 | 1280x720 | Scratches, Cracks, Wear | Collected from production line |
| Gears (Type B) | 1000 | 1280x720 | Broken Teeth, Burrs, Wear | Collected from production line |
| Bolts (Type C) | 800 | 1280x720 | Deformation, Cracks, Rust | Laboratory simulations |
| Mixed Defects Set | 700 | 1280x720 | Various Defects | Combination of different parts and defect types |

The image dataset was split into training and testing sets with an 80% to 20% ratio. We ensured an even distribution of different parts and their defect types across both sets to prevent any imbalance in model performance on certain defect types. The data preprocessing stage included standardization of all images, such as resizing and grayscaling. Additionally, to enhance the model's robustness, various augmentation techniques were applied to the images, including rotation, scaling, translation, mirroring, and color jittering[15]. These preprocessing steps not only increase data diversity but also enable the model to better adapt to varying lighting conditions and environmental changes. For training the YOLO model, image annotation is an essential step. We used professional annotation tools to manually mark the defect areas in all images, ensuring precision and consistency through a rigorous review process[16]. The defect areas in each image were marked with rectangular bounding boxes, along with detailed annotations of defect type and location. During annotation, we adopted multiple labeling strategies to optimize the model's detection and classification performance, including: Severity-based labeling: Defects were categorized into minor, moderate, and severe based on their intensity. Type-based labeling: Different defect types, such as cracks, wear, deformation, etc., were categorized. Type labeling without RoI distinction: In cases where the region of interest (RoI) could not be clearly distinguished, the entire image was labeled uniformly for defect type. These annotations were converted into a format required by the YOLO model, including each defect's category, bounding box coordinates, and normalized values relative to the image size. Through this detailed annotation method, the model could better learn the characteristics of each defect during training, thereby improving its detection and classification accuracy[17].

### 3.2. Improved YOLO Algorithm Model and Principles

In the practical application of product defect detection, although the original YOLO algorithm has efficient detection speed and high accuracy, it still has certain limitations when facing complex backgrounds and multi-scale targets[18]. Therefore, in this study, we improved the YOLO algorithm by introducing deeper network structures and optimizing the feature extraction modules, enhancing the model's performance in detecting machine part defects[19].

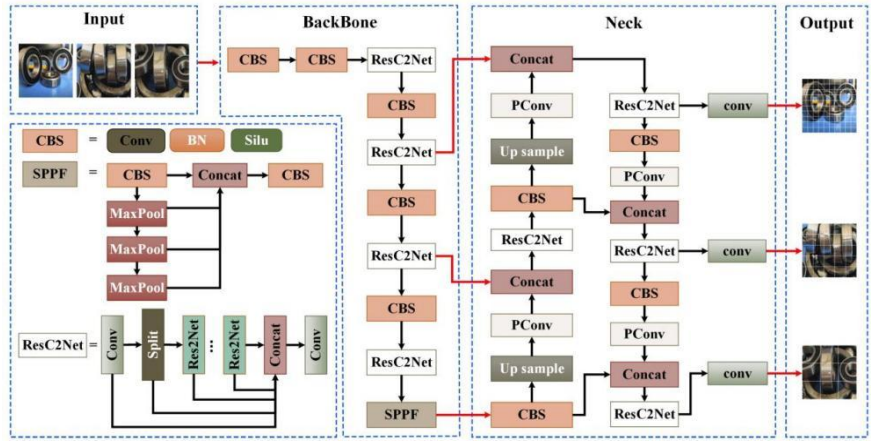

*Figure 2: Improved YOLO Algorithm Model Architecture*

<Figure 2> shows the architecture of the improved YOLO algorithm model, which is mainly composed of three parts: BackBone (main network), Neck (neck structure), and Output (output layer). Throughout the model, we optimized and adjusted various parts of the original YOLO to different extents. The BackBone network part adopts an improved ResC2Net structure, with its core idea being the enhancement of multi-scale feature expression through the introduction of residual branches. In the ResC2Net module, the input image first undergoes convolution, and then the feature map is split into multiple sub-feature maps, which are processed separately by multiple residual networks (Res2Net), and finally, the processed sub-feature maps are reassembled through concatenation. Mathematically, the ResC2Net module can be expressed as shown in Equation 1:

$$F_{ResC2Net}(x) = Concat(f_1(x), f_2(x), ..., f_3(x))$$

where $f_i(x)$ represents the processing result of the i-th sub-feature map, and n is the number of sub-feature maps. Through this improvement, the model can effectively capture features of different scales, thereby enhancing its detection capability in complex scenarios. In the Neck part, the model further improves detection accuracy by introducing a multi-level feature fusion strategy. Specifically, we used cross-layer connections and upsampling operations to allow features of different scales to fuse. To better retain key information, we also introduced PConv (Partial Convolution) operations, which effectively reduce the number of parameters and improve the mode's computational efficiency. Additionally, the SPPF (Spatial Pyramid Pooling-Fast) module was introduced into the Neck part to extract global contextual information by pooling features at different scales. The SPPF module can be expressed as shown in Equation 2:

$$F_{SPPF}(x) = Concat(MaxPool_1(x), MaxPool_2(x), ..., MaxPool_n(x))$$

where $MaxPool_i(x)$ represents the output of the i-th pooling operation. The SPPF module enriches the global context of the features by applying maximum pooling at different scales, thereby improving the model's ability to detect small targets and complex background areas. In the output layer, the improved YOLO model adopts a multi-scale prediction strategy. The model performs convolution operations on feature maps of different resolutions, generating prediction boxes of different scales to accurately detect multi-scale targets. This strategy not only improves the model's detection accuracy but also effectively reduces the miss rate. Through these improvements, the YOLO model's performance in detecting machine part defects has been significantly enhanced. The model can accurately detect various types of defects and maintain high detection accuracy in complex backgrounds and multi-scale targets. Further experiments have shown that the improved YOLO model has great potential for practical application in industrial production environments, providing an effective solution for the automated quality inspection of machine parts[20].

*3.3. Network Training and Optimization*

After preparing the dataset and designing the model architecture, network training and optimization are critical steps to achieving efficient machine part defect detection. To ensure that the improved YOLO model performs well in complex industrial environments, we meticulously designed the training process, including hyperparameter tuning, application of data augmentation techniques, optimizer selection, loss function design, and training strategy adjustments. These strategies not only improved the model's detection accuracy but also enhanced its generalization ability[21].Firstly, hyperparameter settings directly impact the training outcome. We used grid search methods to fine-tune key hyperparameters. The final settings included a learning rate of 0.001, a batch size of 16, and a weight decay coefficient of 0.0005. These hyperparameters were adjusted based on the model's performance during the early training stages to ensure fast convergence and avoid local minima. The learning rate adjustment strategy adopted a stepwise decay, where the learning rate was gradually reduced when the loss on the validation set no longer significantly decreased, helping the model fine-tune better in the later stages of training[22]. Data augmentation is an important means of preventing overfitting and improving generalization. During training, we applied various augmentation operations to the training dataset, such as random rotation, scaling, translation, color jittering, and random cropping. These operations aim to increase data diversity, allowing the model to better adapt to different lighting conditions and background variations. The data augmentation process significantly enriched the diversity of the training samples, thereby enhancing the model's performance when faced with real-world complex scenarios[23]. The choice of optimizer is crucial to the training efficiency and final performance of the model. We chose the Adam optimizer, which combines the advantages of momentum and adaptive learning rate adjustments, allowing the model to dynamically adjust the learning rate under different gradient conditions, thus speeding up convergence and improving stability. The core formula of the Adam optimizer is as shown in Equation 3,4,5:

$$m_t = \beta_1 m_{t-1} + (1 - \beta_1)g_t$$

$$v_t = \beta_2 v_{t-1} + (1 - \beta_2)g_t^2$$

$$\theta_t = \theta_{t-1} - \frac{\alpha \widehat{m_t}}{\sqrt{\widehat{v_t}} + \epsilon}$$

where $m_t$ and $v_t$ represent the first-order and second-order momentum estimates of the gradient, respectively, $\alpha$ is the learning rate, and $\epsilon$ is a small constant to prevent division by zero. Through these formulas, the Adam optimizer effectively adjusts the update magnitude of each parameter, making the model more adaptive during training. The design of the loss function is central to guiding model learning. In the YOLO model, the loss function consists of localization loss, classification loss, and confidence loss. To improve the model's detection accuracy, we introduced Intersection over Union (IoU) as a metric in the localization loss, as shown in Equation 6:

$$\text{IoU} = \frac{\text{Area of Overlap}}{\text{Area of Union}}$$

By introducing IoU, we can more precisely measure the overlap between the predicted bounding box and the actual bounding box, thereby optimizing the model's localization capability. Additionally, to balance the influence of positive and negative samples during training, we applied weighted handling in the confidence loss section, ensuring that the model is not distracted by negative samples when detecting rare defect types. The model was trained on an NVIDIA Tesla V100 GPU to accelerate computation. The entire training process lasted for 200 epochs, with testing on the validation set conducted at the end of each epoch to monitor loss changes and adjust the learning rate as needed. We also employed an early stopping strategy, terminating training early when the validation set performance no longer improved to avoid overfitting[24]. After the training and optimization process, the model exhibited excellent performance on the test set. The test results indicated that the improved YOLO model not only accurately identifies various types of machine part defects but also maintains high detection accuracy and efficiency in complex backgrounds and multi-scale targets[25].

## 4. Experiments and Results Analysis

To validate the performance of the improved YOLO model in detecting defects in machine parts, we designed and conducted a series of rigorous experiments. These experiments not only evaluated the model's detection precision, recall, and mean average precision (mAP) but also examined its robustness and real-time performance in different complex scenarios. The experimental data were derived from the previously prepared datasets, including various types of machine parts and their common defects. The experiments were divided into three main steps: model training, performance evaluation, and result analysis. The dataset used in the experiments covers multiple categories of machine parts, such as bearings, gears, and bolts. The quantity of images and types of defects for each category were detailed in the dataset preparation section[27]. To comprehensively evaluate the model's performance, we split the dataset into 80% for training and 20% for testing, ensuring that the distribution of defect types within each subset was balanced. In the experiments, all model training was conducted on an NVIDIA Tesla V100 GPU. During training, the Adam optimizer was used, with an initial learning rate of 0.001, dynamically adjusted based on the loss curve. The model was trained for a total of 200 epochs, with performance evaluations conducted on the validation set at the end of each epoch to monitor the model's convergence and avoid overfitting[28]. The model's performance was assessed using precision, recall, F1 score, and mean average precision (mAP) metrics. <Table 2> shows the detection results for different defect types on the test set. All metrics were calculated at a standard threshold of 0.5.

*Table 2: Performance Evaluation of the Model on the Test Set*

| Defect Type | Precision | Recall | F1 Score | mAP |
|---|---|---|---|---|
| Bearing - Scratch | 0.95 | 0.93 | 0.94 | 0.92 |
| Bearing - Crack | 0.94 | 0.91 | 0.92 | 0.91 |
| Gear - Broken Teeth | 0.96 | 0.94 | 0.95 | 0.93 |
| Bolt - Deformation | 0.92 | 0.89 | 0.91 | 0.89 |
| Bolt - Rust | 0.93 | 0.90 | 0.91 | 0.90 |

| Overall Defect Detection | 0.94 | 0.92 | 0.93 | 0.91 |

The experimental results indicate that the improved YOLO model demonstrates high precision and recall across various defect detection tasks. Among them, the scratch detection accuracy for bearings was the highest, reaching 0.95, while the performance for detecting bolt deformations was relatively weaker but still maintained a high level at 0.92. The overall mean average precision (mAP) remained stable above 0.91, indicating that the model's detection capabilities were fairly balanced across different defect types.

*Table 3: Real-Time Performance Analysis of the Model in Different Scenarios*

| Scenario Type | Input Image Size | Average Detection Time (ms) | Frame Rate (FPS) |
| --- | --- | --- | --- |
| Simple Background | 1280x720 | 25 | 40 |
| Complex Background | 1280x720 | 30 | 33 |
| Multi-Target Detection | 1280x720 | 32 | 31 |
| High-Resolution Image | 1920x1080 | 50 | 20 |

<Table 3> shows the detection speed of the model in scenarios of varying complexity. The results indicate that the model handles simple backgrounds and multi-target detection tasks with average detection times of 25ms and 32ms, respectively, and frame rates (FPS) between 31 and 40, which meets the requirements for real-time detection. However, when processing high-resolution images, the average detection time increased to 50ms, and the frame rate decreased to 20 FPS, suggesting that while the model consumes more computational resources when handling high-resolution images, it still maintains relatively smooth detection performance. Through the analysis of the experimental results, it is evident that the improved YOLO model performs well in detecting defects in machine parts[29]. The model consistently maintains high detection precision and recall across various defect types, particularly in complex backgrounds where it accurately detects even minor defects[30]. This success can be attributed to the enhanced multi-scale feature capture capabilities of the improved ResC2Net structure and SPPF module. Additionally, the model shows excellent real-time performance, achieving fast detection in various scenarios, especially in production line settings where real-time detection is critical[31]. This provides technical support for practical industrial applications, ensuring efficient defect detection without compromising production efficiency. However, the experimental results also highlight that the model's computational resource consumption increases significantly when processing high-resolution images, leading to a decrease in detection speed. Therefore, in practical applications, there is a need to balance detection precision and real-time performance by adjusting the input image resolution and detection strategy according to specific needs to ensure optimal model performance in particular scenarios[32].

## 5. Applications and Discussion

The improved YOLO model developed in this study has demonstrated exceptional performance in detecting defects in machine parts, excelling in both detection precision and recall. Moreover, it maintains high real-time performance even in complex scenarios, making it highly suitable for practical industrial applications, particularly in quality control and automated inspection in manufacturing. By accurately detecting defects in various types of machine parts, the model can promptly identify potential issues during production, reducing the incidence of defective products and enhancing overall production quality[33]. In practical applications, the improved YOLO model can be integrated into real-time detection systems on production lines, serving as a core component of quality control[34]. Its rapid detection capabilities can help manufacturing enterprises quickly identify and label defective parts as products pass through the production line, enabling subsequent processes to address these issues more efficiently. Additionally, the model's multi-scale feature detection capability allows it to handle a wide range of complex industrial scenarios, whether it is detecting single defects in simple backgrounds or multiple defects in complex backgrounds, the model exhibits high robustness[35]. However, despite the significant improvements in the model's performance, there are still some challenges and limitations in practical applications[36]. First, while the model performs well in detecting defects in complex backgrounds and high-resolution images, this comes at the cost of significantly increased computational resource consumption. The decrease in detection speed when processing high-resolution images may impact scenarios that demand extremely high real-time performance. Therefore, when deploying the model in practice, it is essential for companies to find the optimal balance between detection precision and speed or alleviate this issue through hardware acceleration and distributed computing[37]. Additionally, the diversity and quality of the dataset are crucial to the model's performance. The image dataset used in this study covers many common machine parts and defects, but in actual industrial environments, a wider variety of parts and defect patterns may emerge[38].

## 6. Conclusion

This study successfully improved the YOLO model for efficient detection of defects in machine parts. Experimental results demonstrated that the enhanced model maintains high accuracy and good real-time performance even in complex scenarios, showing strong potential for application in industrial production environments. Although there is some performance trade-off when handling high-resolution images, the model can effectively balance detection precision and speed through appropriate configuration and optimization. Future research could further optimize the model structure and expand the dataset to meet broader industrial detection needs. Overall, the improved YOLO model offers a viable technical pathway for enhancing quality control in manufacturing.[39]

**References**

[1] Li, Zhenglin, et al. "Stock market analysis and prediction using LSTM: A case study on technology stocks." Innovations in Applied Engineering and Technology (2023): 1-6.

[2] Mo, Yuhong, et al. "Large Language Model (LLM) AI Text Generation Detection based on Transformer Deep Learning Algorithm." International Journal of Engineering and Management Research 14.2 (2024): 154-159.

[3] Li, Shaojie, Yuhong Mo, and Zhenglin Li. "Automated pneumonia detection in chest x-ray images using deep learning model." Innovations in Applied Engineering and Technology (2022): 1-6.

[4] Mo, Yuhong, et al. "Password complexity prediction based on roberta algorithm." Applied Science and Engineering Journal for Advanced Research 3.3 (2024): 1-5.

[5] Song, Jintong, et al. "A comprehensive evaluation and comparison of enhanced learning methods." Academic Journal of Science and Technology 10.3 (2024): 167-171.

[6] Liu, Tianrui, et al. "Spam detection and classification based on distilbert deep learning algorithm." Applied Science and Engineering Journal for Advanced Research 3.3 (2024): 6-10.

[7] Qi Z, Ma D, Xu J, et al. Improved YOLOv5 Based on Attention Mechanism and FasterNet for Foreign Object Detection on Railway and Airway tracks[J]. arXiv preprint arXiv:2403.08499, 2024.

[8] Ma, Danqing, et al. "Comparative analysis of x-ray image classification of pneumonia based on deep learning algorithm algorithm." Research Gate 8 (2024).

[9] Xiang A, Huang B, Guo X, et al. A neural matrix decomposition recommender system model based on the multimodal large language model[J]. arXiv preprint arXiv:2407.08942, 2024.

[10] Ma D, Wang M, Xiang A, et al. Transformer-Based Classification Outcome Prediction for Multimodal Stroke Treatment[J]. arXiv preprint arXiv:2404.12634, 2024.

[11] Xiang A, Qi Z, Wang H, et al. A Multimodal Fusion Network For Student Emotion Recognition Based on Transformer and Tensor Product[J]. arXiv preprint arXiv:2403.08511, 2024.

[12] Dai, Shuying, et al. "The cloud-based design of unmanned constant temperature food delivery trolley in the context of artificial intelligence." Journal of Computer Technology and Applied Mathematics 1.1 (2024): 6-12.

[13] Mo, Yuhong, et al. "Make Scale Invariant Feature Transform "Fly" with CUDA." International Journal of Engineering and Management Research 14.3 (2024): 38-45.

[14] He, Shuyao, et al. "Lidar and Monocular Sensor Fusion Depth Estimation." Applied Science and Engineering Journal for Advanced Research 3.3 (2024): 20-26.

[15] Liu, Jihang, et al. "Unraveling large language models: From evolution to ethical implications-introduction to large language models." World Scientific Research Journal 10.5 (2024): 97-102.

[16] Mo, Yuhong & Zhang, Yuchen & Li, Hanzhe & Wang, Han & Yan, Xu. (2024). Prediction of heart failure patients based on multiple machine learning algorithms. Applied and Computational Engineering. 75. 1-7. 10.54254/2755-2721/75/20240498.

[17] Tang X, Wang Z, Cai X, et al. Research on Heterogeneous Computation Resource Allocation based on Data-driven Method[J]. arXiv preprint arXiv:2408.05671, 2024.

[18] Yukun, Song. "Deep Learning Applications in the Medical Image Recognition." American Journal of Computer Science and Technology 9.1 (2019): 22-26.

[19] Song, Yukun, et al. "Going Blank Comfortably: Positioning Monocular Head-Worn Displays When They are Inactive." Proceedings of the 2023 ACM International Symposium on Wearable Computers. 2023.

[20] Song, Yukun, et al. "Looking From a Different Angle: Placing Head-Worn Displays Near the Nose." Proceedings of the Augmented Humans International Conference 2024. 2024.

[21] Zhu W, Hu T. Twitter Sentiment analysis of covid vaccines[C]//2021 5th International Conference on Artificial Intelligence and Virtual Reality (AIVR). 2021: 118-122.

[22] Hu T, Zhu W, Yan Y. Artificial intelligence aspect of transportation analysis using large scale systems[C]//Proceedings of the 2023 6th Artificial Intelligence and Cloud Computing Conference. 2023: 54-59.

[23] Zhu W. Optimizing distributed networking with big data scheduling and cloud computing[C]//International Conference on Cloud Computing, Internet of Things, and Computer Applications (CICA 2022). SPIE, 2022, 12303: 23-28.

[24] Yan Y. (2022). Influencing Factors of Housing Price in New York-analysis: Based on Excel Multi-regression Model. In Proceedings of the International Conference on Big Data Economy and Digital Management - Volume 1: BDEDM, ISBN 978-989-758-593-7, pages 1005-1009. DOI: 10.5220/0011362000003440

[25] Li, Xintao, and Sibei Liu. "Predicting 30-Day Hospital Readmission in Medicare Patients: Insights from an LSTM Deep Learning Model." medRxiv (2024): 2024-09.

[26] Yan, Hao, et al. "Research on image generation optimization based deep learning." Proceedings of the International Conference on Machine Learning, Pattern Recognition and Automation Engineering. 2024.

[27] Li K, Xirui P, Song J, et al. The application of augmented reality (ar) in remote work and education[J]. arXiv preprint arXiv:2404.10579, 2024.

[28] Li K, Zhu A, Zhou W, et al. Utilizing deep learning to optimize software development processes[J]. arXiv preprint arXiv:2404.13630, 2024.

[29] Mo, Kangtong, et al. "DRAL: Deep Reinforcement Adaptive Learning for Multi-UAVs Navigation in Unknown Indoor Environment." arXiv preprint arXiv:2409.03930 (2024).

[30] Qiao, Yuxin, et al. "Robust Domain Generalization for Multi-modal Object Recognition." arXiv preprint arXiv:2408.05831 (2024).

[31] Li, Keqin, et al. "Optimizing Automated Picking Systems in Warehouse Robots Using Machine Learning." arXiv preprint arXiv:2408.16633 (2024).

[32] Lu, Qingyi, et al. "Research on adaptive algorithm recommendation system based on parallel data mining platform." Advances in Computer, Signals and Systems 8.5 (2024): 23-33.

[33] Wu, Zhizhong, et al. "Research on Prediction Recommendation System Based on Improved Markov Model." Advances in Computer, Signals and Systems 8.5 (2024): 87-97.

[34] Chunyan Mao, Shuaishuai Huang, Mingxiu Sui, Haowei Yang, Xueshe Wang, Analysis and Design of a Personalized Recommendation System Based on a Dynamic User Interest Model. Advances in Computer, Signals and Systems (2024) Vol. 8: 109-118. DOI: http://dx.doi.org/10.23977/acss.2024.080513.

[35] Huang, Shuaishuai, et al. "Deep Adaptive Interest Network: Personalized Recommendation with Context-Aware Learning." arXiv preprint arXiv:2409.02425 (2024).

[36] Wang, Xueshe. Nonlinear Energy Harvesting with Tools from Machine Learning. Diss. Duke University, 2020.

[37] Wang, Xue-She, et al. "A model-free sampling method for basins of attraction using hybrid active learning (HAL)." Communications in Nonlinear Science and Numerical Simulation 112 (2022): 106551.

[38] Wang, Xue-She, James D. Turner, and Brian P. Mann. "Constrained attractor selection using deep reinforcement learning." Journal of Vibration and Control 27.5-6 (2021): 502-514.

[39] Wang, Xue-She, and Brian P. Mann. "Attractor selection in nonlinear energy harvesting using deep reinforcement learning." arXiv preprint arXiv:2010.01255 (2020).